# SEMANTICS-BASED SERVICES FOR A LOW CARBON SOCIETY: AN APPLICATION ON EMISSIONS TRADING SYSTEM DATA AND SCENARIOS MANAGEMENT

CECILIA CAMPOREALE, ANTONIO DE NICOLA, MARIA LUISA VILLANI

Computing and Technological Infrastructures Lab,
Technical Unit for Energy and Environmental Modelling (UTMEA)
Italian National Agency for New Technologies, Energy and Sustainable Economic Development (ENEA)
Casaccia Research Centre, Via Anguillarese 301, 00123 Roma, Italy
cecilia.camporeale@enea.it, antonio.denicola@enea.it, marialuisa.villani@enea.it

**Abstract**
A low carbon society aims at fighting global warming by stimulating synergic efforts from governments, industry and scientific communities. Decision support systems should be adopted to provide policy makers with possible scenarios, options for prompt countermeasures in case of side effects on environment, economy and society due to low carbon society policies, and also options for information management. A necessary precondition to fulfill this agenda is to face the complexity of this multi-disciplinary domain and to reach a common understanding on it as a formal specification. Ontologies are widely accepted means to share knowledge. Together with semantic rules, they enable advanced semantic services to manage knowledge in a smarter way. Here we address the European Emissions Trading System (EU-ETS) and we present a knowledge base consisting of the EREON ontology and a catalogue of rules. Then we describe two innovative semantic services to manage ETS data and information on ETS scenarios.

**Keywords:** Low Carbon Society, Emissions Trading System, Ontology, Semantic Rule, Semantics-based Technology

## 1. Introduction

This work focuses on the European Emissions Trading System (EU-ETS) [WR1], one of the major pillars of the European strategy to support a low-carbon society (LCS). The LCS is defined as a "*society that should take actions that are compatible with the principles of sustainable development, ensuring that the development needs of all groups within society are met; make an equitable contribution towards the global effort to stabilize the atmospheric concentration of $CO_2$ and other greenhouse gases at a level that will avoid dangerous climate change, through deep cuts in global emissions; demonstrate a high level of energy efficiency and use low-carbon energy sources and production technologies; and adopt patterns of consumption and behavior that are consistent with low levels of greenhouse gas emissions*" [Skea, Nishioka, 2008].

A LCS strategy can be supported by policies. Besides environment, such policies could have impact on economy, industry, energy system and, in general, on the way of living of citizens (i.e., societal system). Thus, before defining new policies, policy makers need to define different scenarios in order to take aware decisions. "*Scenarios are possible future states of the world representing plausible conditions under different assumptions*" [Mahmoud et al., 2009]. Their development requires taking into account large datasets concerning different domains, and different countries. Managing such complexity requires tackling several issues. Among them, we cite those covered in this work:

1. **need for an integrated environmental modeling (IEM) approach**. LCS concerns multi-disciplinary and inter-dependent knowledge. IEM is considered as one of the most promising approaches to define break-through solutions leveraging on such knowledge [Laniak et al., 2013] [Argent, 2004].
2. **sharing common understanding**. A complete scenario (made by the mix-up of energetic, industrial, technological, political and environmental aspects) should be framed into a shareable "scientific" formulation. The importance of involving a multi-disciplinary community of people [Laniak et al., 2013] [Krueger et al., 2012] is a precondition to reach such understanding.
3. **structural and behavioural knowledge modelling**. The LCS domain concerns both the structural aspects of entities (e.g., the energy price) and others related to how entities behave or should behave (e.g., the allocated $CO_2$ emissions of the ACME company are $37 \bullet 10^4$ $tCO_2$). A modelling approach covering both aspects is required.
4. **managing large governmental datasets**. LCS data are needed for LCS analysis and definition of scenarios. Although their *availability* from governments might be an issue [Zicari, 2013] due to the lack of competitive pressure for making data open, one problem is to assess their *quality* and *veracity* since such data are usually manually collected by error-prone processes. Further relevant problems related to data management concern data integration and data privacy.
5. **need to develop multi-domain scenarios**. In order to use "federated" scenarios, all the problem's components should be made interoperable and fed into a common scheme.

The present work proposes a semantics-based approach to address these issues. Indeed, semantic technologies based on ontologies help to better organize data [Poggi et al., 2008] [Reichman et al., 2011] and enable development of intelligent services for data management (e.g., to verify LCS data quality and to better share and query them). An ontology is a formal specification of a shared conceptualization [Borst, 1997] [Gruber, 1993]. It consists of a set of concepts, a set of relationships between them, and a set of axioms. In this respect, an ontology is a means for achieving a common and comprehensive representation of all the elements in the complex LCS problem space. Also, an ontology may constitute the basis for the design of a decision support system enabling decision makers, from civil servants to politicians, to manage such a large domain through a *what-if* analysis and optimized decision-making activities with respect to various adoptable policies.

Semantic technologies applied to environmental domains are gathering a lot of interest from the research community [Villa et al., 2008]. Here, the idea of using semantic technologies for LCS has been applied to the EU-ETS [WR1]. Essentially, the EU-ETS is a mechanism to reduce greenhouse gases emissions in Europe through a market of emissions allowances involving energy-intensive industries of the EU countries. Around 45% of EU emissions are limited by the EU-ETS, and, by acting on the carbon price, investments in low-carbon technologies are promoted. To this aim, the EU-ETS leverages on a data collection and monitoring system of emissions by energy-intensive industries of the EU countries.

More generally, the EU-ETS is an instrument of political management involving industrial strategies, technological platforms, political and social issues, and environmental needs. It represents the point of contact of the environmental and industrial policies and, as such, it should contain measures able to comply with and harmonize country's environmental and industrial policies.

Because of this tight entanglement of different issues, the ETS research field must be tackled by multidisciplinary groups where diverse competences must collaborate. Environmental analysis should produce input data for defining technological benchmarks to be fulfilled, in order to comply with required milestones (in terms of overall emissions).

Thus, the quality of the ETS data to base these analyses, and integration and sharing of data from multi-disciplinary studies, are major issues for a EU-ETS data management system, which can be faced through semantics-based techniques.

In this setting, we have defined a semantics-based architecture and two innovative semantic services to manage ETS data and ETS scenarios' information. The knowledge base consists of the EREON (Ets-Related European ONtology) and of a catalogue of rules. The ontology is the common grounding for the collaborative work of the scientists and enables holistic studies. The rules allow modeling constraints, policies and expected behaviors. As stated in [Laniak et al., 2013], managing information related to multi-disciplinary domains is a challenge since semantic consistency should be guaranteed when integrating models (or ontologies) from different fields. Our proposal addresses such issue by means of a rules-based approach, where the rules are used to define mapping relationships between the conceptual representations of data and to represent data constraints.

This work is addressed to Governments (National & EU), in particular, *policy makers* who take strategic decisions about the ETS and need tools to evaluate the impact of their strategies; *LCS researchers and analysts*, who need to collaborate on the construction of shared knowledge about ETS as a common basis upon which to perform their studies and integrate their results; *industrial stakeholders*, who need to better manage their own ETS-related data to be compliant with governmental objectives; and *knowledge engineers*, who want to try new applications for semantic languages and tools, and search for new challenges.

It should be noted that our proposal, in developing the semantic services, takes into account the *data dogmatism* issue, i.e., the complete trust in data [Zicari, 2013]. In particular, we deem that data analysis can provide quite remarkable insights, but we are aware that domain experts and common sense must continue to play a role in the final decisions.

This work originated from a scientific collaboration running between ENEA, the Italian National Agency for New Technologies, Energy and Sustainable Economic Development, and the Italian Ministry of Economic Development [WR2]. Preliminary results were presented at two workshops ([Ciorba et al., 2013], [Camporeale et al., 2013]). In particular, the first paper describes the upper level ontology designed as a foundation for ETS-related ontologies. The second paper, in turn, presents EREON, focusing on the ETS context representation and related semantic rules. Furthermore, a service-based architecture built on top of it is presented, with the specification of two useful services, the former for data acquisition and the latter for data analysis. Here we provide the overall picture of our research concerning how semantics-based technologies can provide benefits to ETS and, in general, to LCS. Additionally, we have defined some guidelines to create new semantic services within our architecture by leveraging on a domain-specific rules taxonomy. Furthermore, we provide a detailed description of the experimentation we have done to assess both the efficacy of semantic techniques and their usability for the given domain.

The paper is organized as follows. Section 2 describes related work in the area. Section 3 describes the ETS domain and the specific ICT requirements for the ETS knowledge representation and analysis. Section 4 presents the semantics-based framework together with the ETS knowledge base, the knowledge base engineering approach adopted, the service-based architecture for ETS data management and how the framework can be applied to solve real world problems. Section 5 provides insights on the EREON ontology. Then Section 6 presents the specification of the semantic services together with the rules' taxonomy and the above-mentioned guidelines. Evaluation of the approach of rules and semantics combined is described in Section 7. Finally, Section 8 provides conclusions.

## 2. State of the Art

Being a recent field, the ETS domain is providing room for new ICT applications. In [Chappin, Dijkema, 2009], an agent-based simulation model is presented to analyze the impact of ETS on the $CO_2$ emissions by power generation companies. The model is implemented through a layered software system where agents specifications are ontology-based, coded in Java and with Repast [Collier, 2001] as run-time platform. However, in this work, formal knowledge representation of the ETS domain is not addressed.

Other works dealing with some of our issues as must be searched within the wider domain of sustainability science. This discipline, introduced at the beginning of this century, requires contributions of scientists from fields of the natural and socio-economic sciences. In this setting, a structured knowledge base encompassing the relevant concepts, problems and findings may represent a common grounding for the collaboration work of the scientists in this multi-disciplinary setting and a valuable source of information for decision makers.

Currently, there are only few efforts devoted to providing a formal specification of knowledge in the scope of sustainability science.

In this line, the work by [Kumazawa et al., 2009], who has pioneered this discipline, proposes a reference model for a knowledge-structuring tool, using ontology engineering methods and technology. Guidelines for developing a sustainability science ontology are given, starting from a basic structure of 5 general concepts such as goal, problem, countermeasure, evaluation, and domain concept, and about 562 specific concepts concerning the sustainability science.

The work [Kraines, Guo, 2011] proposes a knowledge sharing system for sustainability science. These authors have developed a web-based platform, called EKOSS, enabling semantic annotation of scientific documents and data by the domain experts and incorporating a reasoning engine to automatically identify semantic links among the knowledge resources, in order to create a knowledge network. This system is based on an ontology called SCINTENG starting from a set of 4 core concepts, such as activity, event, physical object, and substance or material. From these concepts, they have detailed the concepts belonging to the sustainability science domain and the analysis of low carbon society scenarios (corresponding to the contextual and scenario views of our reference model).

With these works we share the motivation for the knowledge base as a means for collaborative work of the various domain experts and the idea of incremental development of the ontology, starting from a small set of core concepts. In particular, we have realized a domain-specific knowledge base and developed semantics-based services to manage and analyse LCS data. We deem that semantic techniques may add value as the increase of available open governmental data ([Reichman et al., 2011], [Gang-Hoon et al., 2014]) requires coping with the variety and heterogeneity characterizing different sectors by means of integrated approaches [Kelly et al., 2013]. Here, such techniques are applied to the EU ETS-related knowledge, a specific subfield of the LCS.

Differently from the works cited above, our solution aims at enriching existing relational databases with semantics, by linking the database schemas to an ontology and using semantic rules for data validation and query. Indeed, this solution increases systems flexibility when varying the rules to be applied and, by working at logical level, eases migration from a technical solution to another one.

Our aim is to demonstrate the efficacy of our solution by exploring the Italian context. A step forward will be integration with scenarios data to set up the basis for a decision support system for the Italian Ministry of Economic Development.

## 3. The ETS Case Study
### 3.1 The ETS Domain
The EU-ETS is a 'cap-and-trade' market-based instrument involving some energy-intensive industries of the 28 EU countries plus Iceland, Liechtenstein and Norway. The EU-ETS is a major pillar of EU climate policy in which the regulator, through the National Allocation Plans (NAP), fixes an upper limit (cap) to the amount of greenhouse gases (GHG) that the involved sectors can emit. Every year, within the cap, a certain number of allowances for emissions production are granted to the firms, where one "allowance" corresponds to one ton of GHG emissions.

Thus, if a firm has reduced its emissions (e.g., with technological intervention), its unused allowances can be sold to another firm in the EU-ETS (trade) or be kept to cover its future needs. In this way, firms exceeding their original allowances can avoid heavy fines. It should be noted that, by putting a price on each ton of carbon emitted, the EU-ETS is driving investments in low-carbon technologies, as industries are forced to account for the cost of emissions allowances in their businesses.

The EU-ETS was launched at the beginning of 2005; after an initial three-year pilot phase of 'learning by doing', a stricter cap (6.5% lower) was established for the period 2008-2012. The scheme has been recently improved with new rules of allocations (auctions) for the allowances. Participation is mandatory for enterprises belonging to the sectors covered by the EU-ETS, which currently includes about 11,000 heavy energy-consuming installations in power generation and manufacturing [Stephan et al., 2014].

By September 2011, the EU Member States have submitted to the EU Commission a list of installations on their territory, covered by the EU-ETS Directive, along with any free allocation to each of those installations. This submission is referred to as the NIMs (National Implementation Measures) [WR3], and has been performed through a data collection template, made available by the European Commission in all the EU languages, in order to ensure harmonization of the data [EU, 2013]. All the installations operators eligible for free allocation of the allowances were required to fill in such a template. This consists of ten sheets and a summary, to provide details about the installation activities with the corresponding emissions. Specifically, these sheets are concerned with: the installation identification within the industrial system and its organization in sub-installations; energy and process emissions; attribution of some GHG emissions to sub-installations; data on energy input, measurable heat and electricity; data about the products of sub-installations and their relation with the EU benchmarks.

Some more information about the content of the NIMs is given in Subsection 5.1.1 as they are part of our case study. Also, documents about the EU-ETS are provided at the EU Commission web site [WR1].

### 3.2 ICT Requirements for ETS Knowledge Representation and Analysis
In this paper we present how semantic technologies can be the basis for a software platform to support domain experts in the modelling and analysis activities related to the EU-ETS and civil servants (e.g., policy makers) in the development of their policies. To this aim, we have identified three main activities that need to be supported by the platform and some challenges related to them that motivate the definition of the ETS knowledge base and have to be addressed in its design.

Context analysis
1.     *Data integration.* The ETS domain is covered by different disciplines, such as economy, law, environment and energy, each with its own vocabulary of technical terms. These terms have to be related with one another in order to enable coexistence of data from different sources and to support cross-domain studies.

2. *Data consistency.* Domain specific studies carried out with different models and tools have to lead to a coherent overall picture of the results. Indeed, these tools already incorporate a basic set of data underpinning the analyses, and these datasets are not independent from each other. Also, the same data may be present within the tools at various levels of granularity, e.g., the same meaning but different serializations.

3. *Data correctness.* This is the basic requirement for reliability of the analysis activities. The ETS data collection mechanisms do not "automatically" guarantee correctness of submitted data (both in the types and in the values), although a certificatory authority must be indicated.

4. *Data model evolution*. The data model at the basis of analysis activities may change over time, either because refinements are needed or to reflect revisions of regulations. For example, any change in the EU-ETS directive that has an impact on the ETS data representation should be addressed accordingly to maintain the alignment between the ETS knowledge base and the ETS domain.

Policy making

5. *Social policy representation*. Civil servants' policies have to be translated into concrete measures to drive quantitative assessment studies and enable algorithmic solutions to policy analysis and decision making. Indeed, political decisions have to be taken by considering their impact on the various governmental objectives. For example, promoting biomass energy consumption as wood burning is attractive as a possible measure for $CO_2$ emissions reduction but this increases the particular matter emissions (PM2.5), with impact on human health.

6. *Collaborative policy assessment*. Civil servants do not necessarily have the technical skills and the complete background knowledge to define models for scenarios development and to quantitatively analyse their policies. On the other hand, the ETS analysts should be informed about the policies that need to be evaluated.

7. *Knowledge building and evolution*. Qualitative analyses of the various policies and their relationships contribute to enrich the knowledge of the field. Making this knowledge explicit and sharing it among the various policy makers is valuable for the formulation of new measures.

Scenarios development

8. *Interoperation for multi-domain analysis and scenarios*. Each of the different fields involved in the ETS domain, such as the environmental, socio-economic and energy fields, has its own established analysis methods and supporting tools. In order to support multi-disciplinary studies, these tools need to interoperate [Kraines et al., 2010]. An interesting example related to the ETS domain could be, for example, the evaluation of the macro-economic effects (such as GDP variation or employment rate) of investments in low carbon technologies in Italy. This study requires scenarios of the Italian energy system, which may be obtained, for example, through the MARKAL-TIMES [WR4] model, jointly with the GAINS [WR5] model, the latter used to describe possible paths of evolution for greenhouse gases (GHGs) emissions. These results could be the basis for macro-economic analyses through methods like Social Accounting Matrix (SAM) [Pyatt, Round, 1977] and GTAP model [Hertel, 1977].

9. *Completeness of the policy assessment*. The combination of the results originated from domain specific scenarios may not result exhaustive as a study of the ETS domain. This provides opportunities for defining new inter-domain relationships and analysis methods.

## 4. The EREON-based Semantic Services (ESS) Framework

The main purpose of EREON-based semantic services is to manage knowledge on ETS and to provide automatic support for the analysis of ETS independently of the particular implementations of the data structures and of the data management technology adopted to this purpose. Essentially, the ESS framework is concerned with ICT systems to be used by:

- *policy makers*, who need automatic support for impact analysis of measures and for decision making;
- *ETS experts*, who need to calibrate their energy/environmental scenario-based models with ETS contextual data to produce long term analysis reports;
- *industrial stakeholders*, who have extended their information systems with environmental data and the data required by the EU-ETS (e.g., the NIMs) to quantify the impact of ETS regulations and political measures over their business, and to regard the emissions trading as a business opportunity [PriceWaterHouseCouper, 2006]. For example, energetic companies, like the Italian power company ENEL, periodically publish reports on environmental aspects, like emissions, fuel consumption, and energy production. So they handle data including those required, for example, by the NIMs [ENEL, 2012].

### 4.1 The ETS Knowledge Base

The core of our proposal for ETS is the ETS knowledge base consisting of EREON, an ontology conceptualizing ETS-related domains, and a set of rules (i.e., ESS rules). We have chosen to use an ontology because it avoids ambiguity problems and also because ontologies are attractive due to the emerging field of cognitive computing where they enable complex decisions-making from a huge amount of information and knowledge [Magazine, A. I. (2010)]. Furthermore, a knowledge base consisting of ontology and rules enables logics-based services, and several reasoners exist, such as RacerPro [WR6], Pellet [WR7], and FaCT++ [Tsarkov, Horrocks, 2006] [WR8] to check knowledge consistency.

### 4.2 ETS Knowledge Base Engineering

The ETS knowledge base has been developed through a collaborative effort of a multi-disciplinary team of 12 domain experts and 3 knowledge engineers with competencies spanning from energy to environment to macroeconomics to ICT. Domain experts provided ontological requirements, terms with their definitions, and constraints. Knowledge engineers have supported them in formalizing concepts and rules through the Protégé ontology management system [WR9] and the available plug in for editing rules. A sketchy representation of the knowledge engineering process is presented in Fig. 1.

Semantic rules (i.e., ESS rules in Fig. 1) attached to the ontology constitute the common basis for various services to enable an effective usage of EREON by the various ETS players (e.g., policy makers, ETS analysts, etc.). The standard languages from the semantic web world have been taken into account for the realization of EREON and of the rules, such as RDF [WR10], OWL [WR11], SWRL [WR12] and/or SPARQL [WR13], to open to a wider collaborative development. Currently, the ontology exists both in OWL Lite, which can be viewed as a profile of OWL 2 [WR14], and in RDF, at the price of losing some expressivity. The RDF version enables integration with relational database management systems as illustrated in the sequel, which is one important advantage of our solution. However, the OWL Lite version of the ontology can in principle be adopted in case a specific service or application requires higher expressivity.

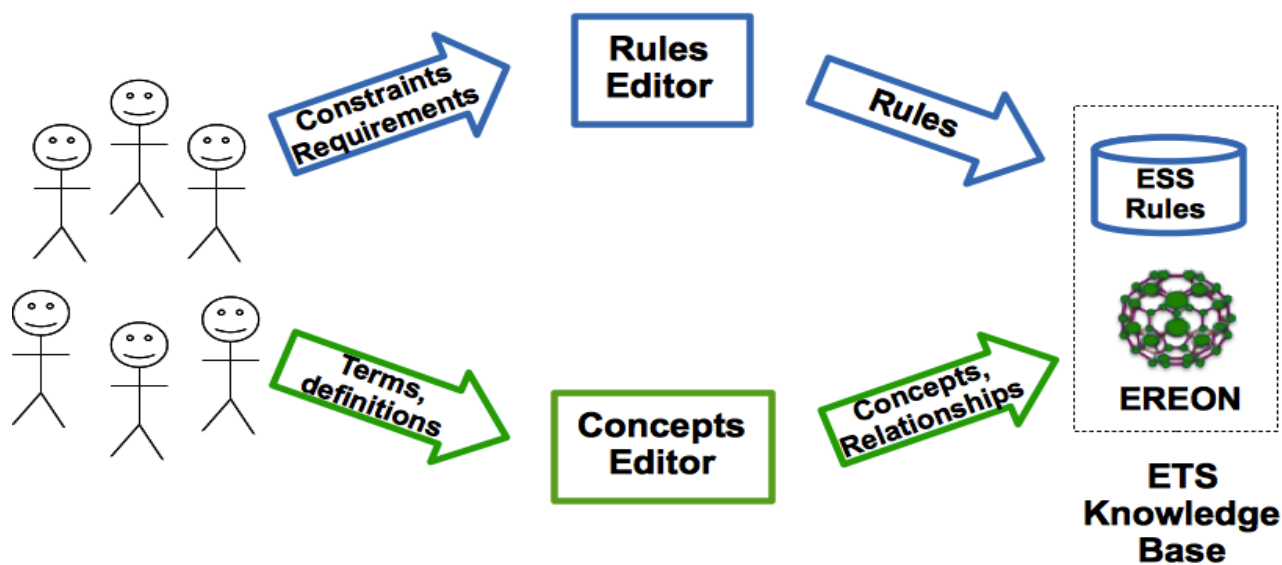

Fig. 1. A sketchy representation of the ETS knowledge base engineering process.

Ontologies are considered "complex artifacts" mainly due to the effort required for their development and also because domain and application competencies and knowledge engineers skills cannot be found easily in the job market. However, currently, there are several methodologies (as [De Nicola et al., 2009], [Fernández-López et al., 1997], and [Sure et al., 2004]) and Natural Language Processing tools (as AlchemyAPI [WR15] and Open Calais [WR16]) supporting modellers in the ontology engineering process.

EREON has been developed through a top-down approach according to the ontology engineering methodologies based on upper-level concepts, such as the ETS ontological foundations [Ciorba et al., 2013]. However, also a bottom-up approach has been used by taking into account existing resources, such as the NIMs [WR3], [De Nicola et al., 2009].

Our idea is that EREON should be an open project to involve in the future a larger community of interested people consisting of domain experts, knowledge engineers, industrial stakeholders, citizens and institutional operators, as main potential users of the ontology. For this reason, we have selected an ontology engineering approach that takes into account collaboration aspects to perform this work. This approach, sketchily represented in Fig. 1, leverages on the works by [De Nicola et al., 2009] and [Barbagallo et al., 2010]. According to [Barbagallo et al., 2010], we may identify two roles in the community of people extending the ontology, i.e., the ontology owner and the ontology participant. The former is responsible for the advancement of the ontology project and for the overall ontology quality. The latter takes part in the project according to her/his skills (e.g., knowledge of ETS domain, knowledge representation competences).

Other non-expert "players" may contribute to the ontology building activity, especially to the policy making part of EREON. For instance, the social participation of industrial enterprises, citizens living near plants and ecological organizations can help in identifying targets and constraints and in proposing measures.

Involvement of non-experts in the ontology engineering process is a relevant challenge currently addressed by the research community [Tempich et al., 2007] [Barbagallo et al., 2010]. Technical difficulties, low participation and lack of coordination in this process are some of the issues that need to be faced. For instance, the paper [Barbagallo et al., 2010] proposes a social approach that uses a voting procedure and natural language processing applications. As stated in [Laniak et al., 2013], the availability of general guidelines, best practices or standards to define and harmonize semantic information is still an open issue. Our ETS knowledge base engineering approach addresses such issue and can be considered as a guideline to develop new semantic contents in other domains related to the low carbon society.

In the following, we mainly focus on the contextual part of the EREON ontology. This has been built together with domain experts and knowledge engineers.

### 4.3 The Service-based Architecture for ETS Data Management

We propose a reference architecture for ETS data management (Fig 2). This comprises EREON, the ESS rules and two semantic services to be used in two different phases: ETS data acquisition and ETS data analysis. This architecture has been defined based on the idea that the semantic services should not replace existing systems managing ETS data (represented in Fig. 2 by a governmental information system), but they should operate on top of these systems in order to provide new and more advanced functionalities. Therefore, the effort required for data replication, for example from a relational DB to RDF triple stores, is avoided. Also, this solution allows us to use new emerging technologies such as D2RQ [WR17], Ontop [WR18] or Virtuoso [WR19], to transform semantic queries into SQL and so moving performance issues to the relational database technology, which is much more consolidated.

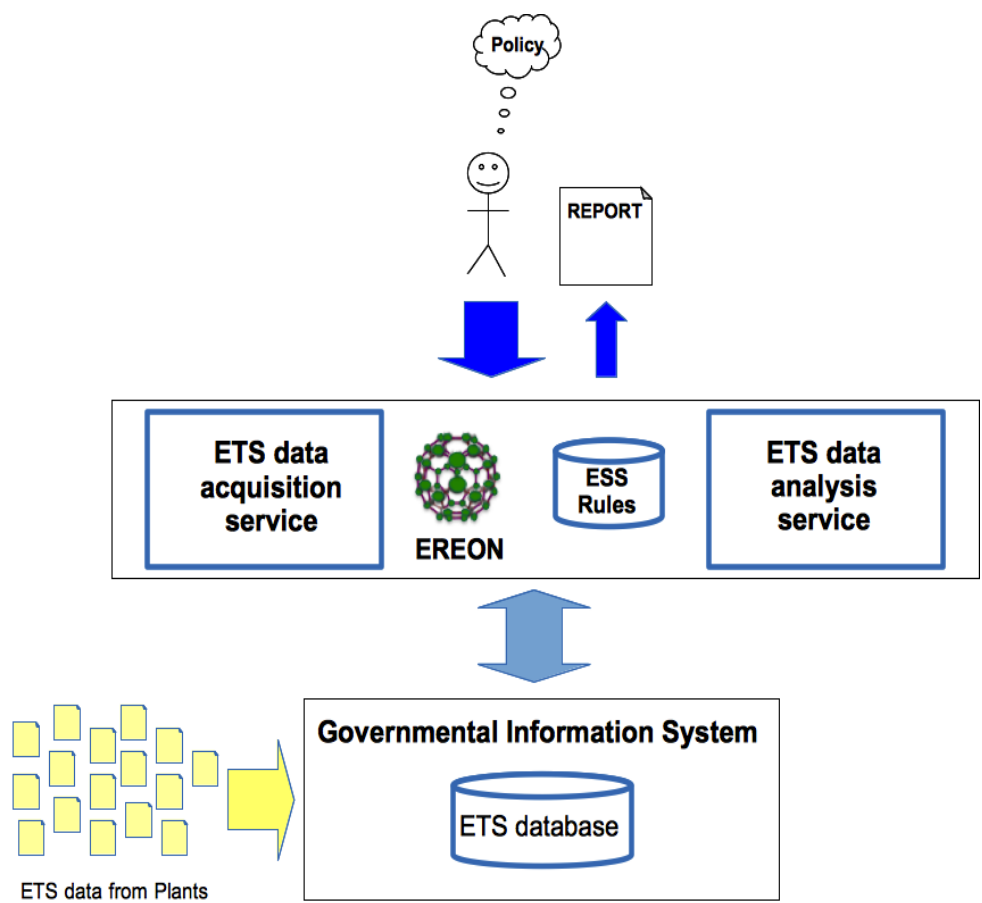

Fig. 2. The ESS reference architecture consisting of EREON, the ESS rules, the ETS data acquisition service and the ETS data analysis service. This architecture is conceived to be built on top of existing systems managed by government (represented in the figure by a Governmental Information System collecting the ETS data from plants).

The requirements that we have elicited for our two services are described in the following.

*Data acquisition phase*
EU governments currently collect information about plants' emissions, energy and electricity consumptions from local plants by means of data model templates common to all EU countries, like the NIMs EXCEL file templates (i.e., NIM's_baseline_data.xls) [WR3]. Although each plant should provide validated documentation signed by a certified authority, such templates are manually filled, which is an error prone process. Once collected from the plants, data are imported in a governmental information system in order to be processed. The data import task is also prone to errors. The aim of the data acquisition phase is to detect rules violations (like potential errors and undesired data) to the purpose of data analysis. Indeed, the rules to be applied in this phase also include user/application dependent rules (e.g., constraints derived from laws) and their violation may be relevant or not according to the analysis one wants to carry out (e.g., missing values from less than 5% of the plants of small size may be not relevant for generating scenarios at National scale).

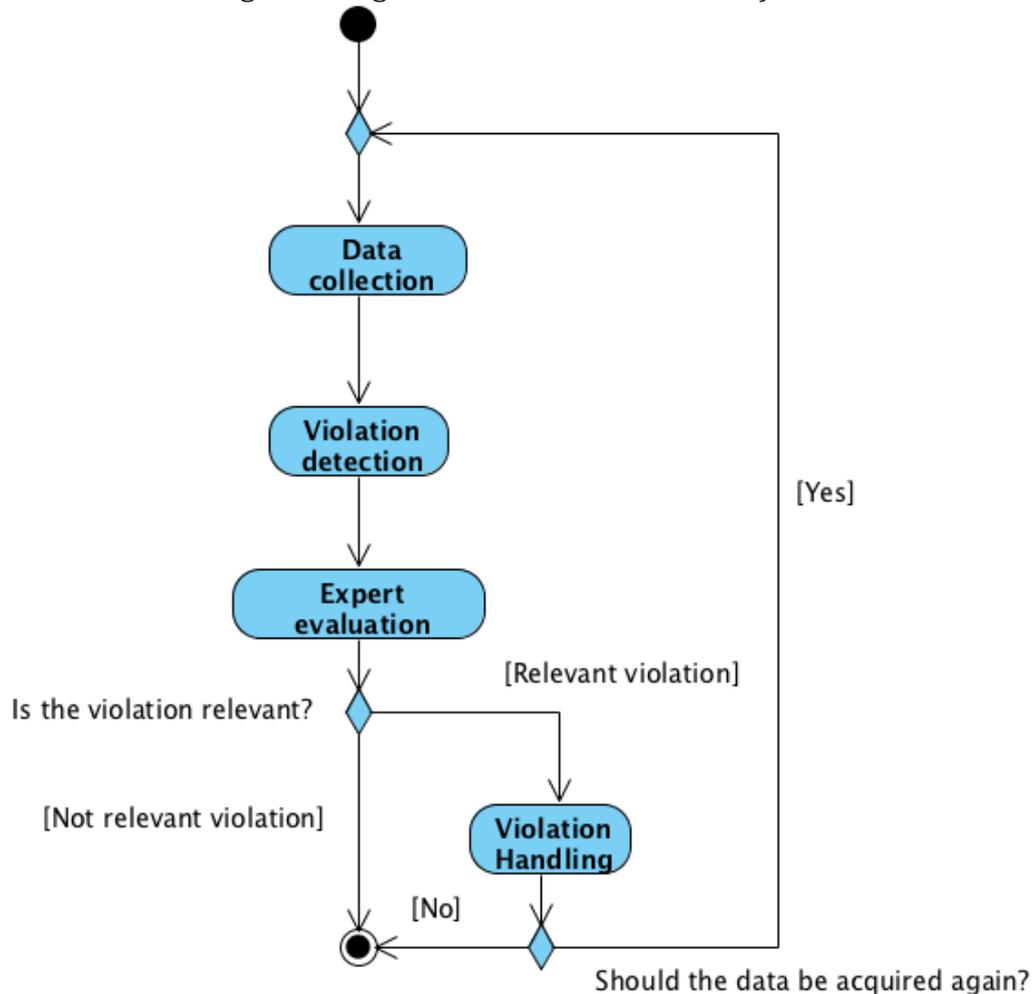

Fig. 3. Data acquisition workflow.

In Fig. 3 we specify a workflow in general terms for this phase, consisting of the following activities:
- **Data collection**. Data produced by the companies subject to the ETS are imported in a governmental information system (see Fig. 2).
- **Violation detection.** A specific set of verification rules are run by the data acquisition service in order to identify violations.
- **Experts evaluation.** Each rule violation identified by the service is evaluated by some experts. This violation can either be judged not relevant or may be considered as an error to be handled by some institutional operator.
- **Violation handling.** A policy for violation handling is put in place by the institutional operator. Actions are decided according to the existing laws. For example, the company may be required to re-submit the data.

*Data analysis phase*
Whenever a policy maker defines a new policy to better manage the ETS, measures envisaged to fulfil such policy (e.g., increasing the $CO_2$ price) could impact on existing plants causing economical losses and social consequences (e.g., increasing unemployment). Here the issue is to detect and quantify the companies impacted by these measures and, consequently, those companies that are indirectly interested in the new policy.

The ETS context data from plants are validated through the ETS data acquisition service, checking the correctness of the information through the verification of a set of semantic rules referring to the EREON ontology. Once imported in the ETS database, the policy maker analyses such data, supported by a set of predefined rules that are executed through the ETS data analysis service. Details about EREON and, specifically, on the rules we have implemented are provided in Section 5 and Section 6.

## 4.4 ESS adoption

The activities of ESS potential users, described at the beginning of Section 4, require the management of large datasets that, usually, are already stored in a specific database and/or distributed in different proprietary analysis tools. Thus, the problem of ESS adoption in wider and existing ICT systems has to be faced. In this case, assuming the existence of a DBMS containing data to be analysed, two types of ESS adoption can be considered:

1. *ESS specification adoption*: the ESS rules are customized according to the data structures at hand, implemented in the query language of the DBMS in place, and executed by the custom application;
2. *ESS implementation adoption:* an ad-hoc mapping between the given database implementation and EREON is provided, to allow ESS rules to be fired by a OWL (or RDF) based rule engine, directly accessing the data stored in the database.

Although the first solution may be more efficient on the performance of the verification and analysis activities, it requires a significant effort in the rules implementation and testing. The second solution, instead, allows taking full advantage of the reasoning capabilities of the semantic technologies but it needs a mapping service between the ontology and the database, as shown in Fig. 4. Therefore we discuss more in details the second solution in the following of the paper.

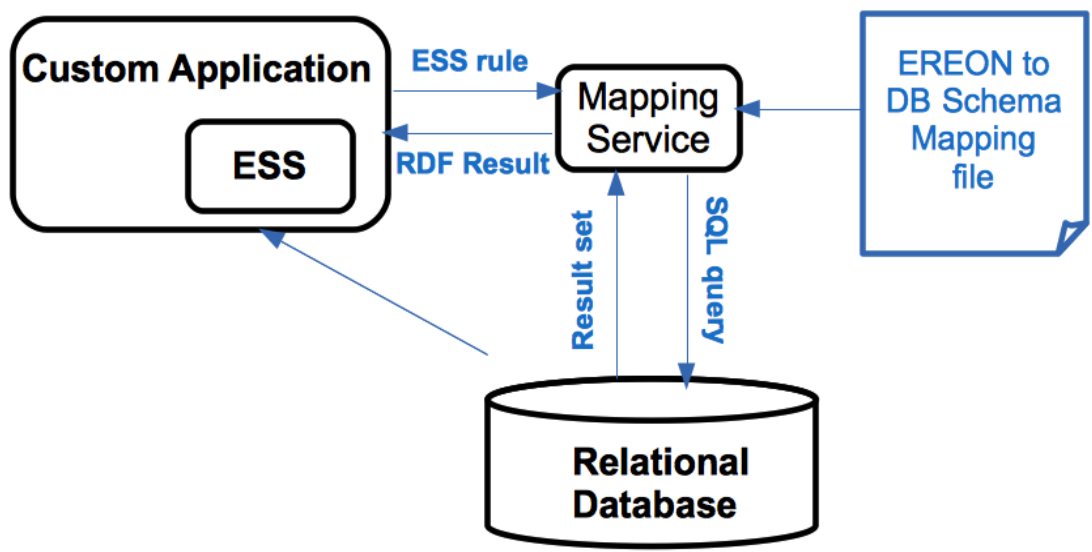

Fig. 4. ESS integration with an existing ICT system.

The Mapping Service is a generic service enabling the execution of the semantic rules by automatically translating them in SQL queries (for a relational database), through a mapping specification between EREON and the particular DB schema. Various technologies are already available to implement the Mapping Service, such as [WR17], [WR18], [WR19] and the only effort for this solution is in the specification of the *EREON to DB schema mapping file*, whose complexity depends on the specific chosen technology. Performance issues for this solution are discussed in Section 7.

## 5. The EREON Ontology
### 5.1 The EREON Ontological Foundations

The ETS-related ontology concerns a multi-disciplinary domain spanning from the societal system to the environment, the production system, the market, the low-carbon technologies, and the energy. This is reflected in EREON's core structure, i.e., the EREON upper level ontology, that consists of a few foundational ETS-related concepts (upper level or meta-concepts) as semantic hooks for the various domains to cover. Starting from these base concepts, an incremental and collaborative ontology building process can be put in place through extension operations, i.e., meronimy (e.g., the chemical sector is part of the Emissions Trading System) and specialization (e.g., the Iron & Steel Industrial Sector is an ETS Sector) relationships.

The foundational concepts have been grouped in three views to address the three categories of requirements we described in Section 3.2. Namely, the *contextual view* is to represent the status of ETS-related entities to be used for collection, integration and sharing of ETS-related data coming from different sources. The *scenario view* is to represent possible temporal evolutions of such entities, which can be obtained by applying scenarios-based models from societal, economic, energetic and environmental studies. The *policy making view* is to represent the policy definition processes and policies evaluation through qualitative and quantitative analyses. It should be noted that these views are not disjoint and, consequently, overlapping exists, consisting in a set of merging concepts (e.g., stakeholder, impact) and connecting together the different scopes of the ontology. For sake of clarity, foundational ETS-related concepts are presented as a UML class diagram in Fig. 5.

The concepts are quite general in their own but they have been linked to represent the EU-ETS domain of interest. Indeed, they have been identified by analysing domain-specific documents, such as the EU-ETS official documents and reports and publications concerning the domain [ENEA, 2013] and produced by ENEA, and by taking into account our application objectives.

The relationships of the upper level concepts can be sustained by the following arguments. The NIMs guidelines [EU, 2013], defined within the EU-ETS, are concerned with some GHG emissions (e.g., $CO_2$, $N_2O$, PFC) by a portion of the *productive system* of the individual State, that have an impact on the *environment* and on the *societal system*. These emissions are caused by the use of some type of *technology* in the industrial processes and $CO_2$ emissions quantities are related to heat and *energy* consumption. The ETS, which is by definition a *market* of emissions allowances, is influenced by, and has an impact on, other subjects (i.e., *stakeholders*) such as governmental institutions, ecological organizations, economic consultancy enterprises, citizens, and finance. Such influence is realized through *policies* and their *measures* (such as energy price variation, technologies incentives [ENEA, 2013]), towards one or more *ETS targets*, such as $CO_2$ emissions reduction. The impact of a measure can be evaluated by producing *scenarios*, following a *computational model* and *forecasting*, *backcasting* or *BAU* techniques.

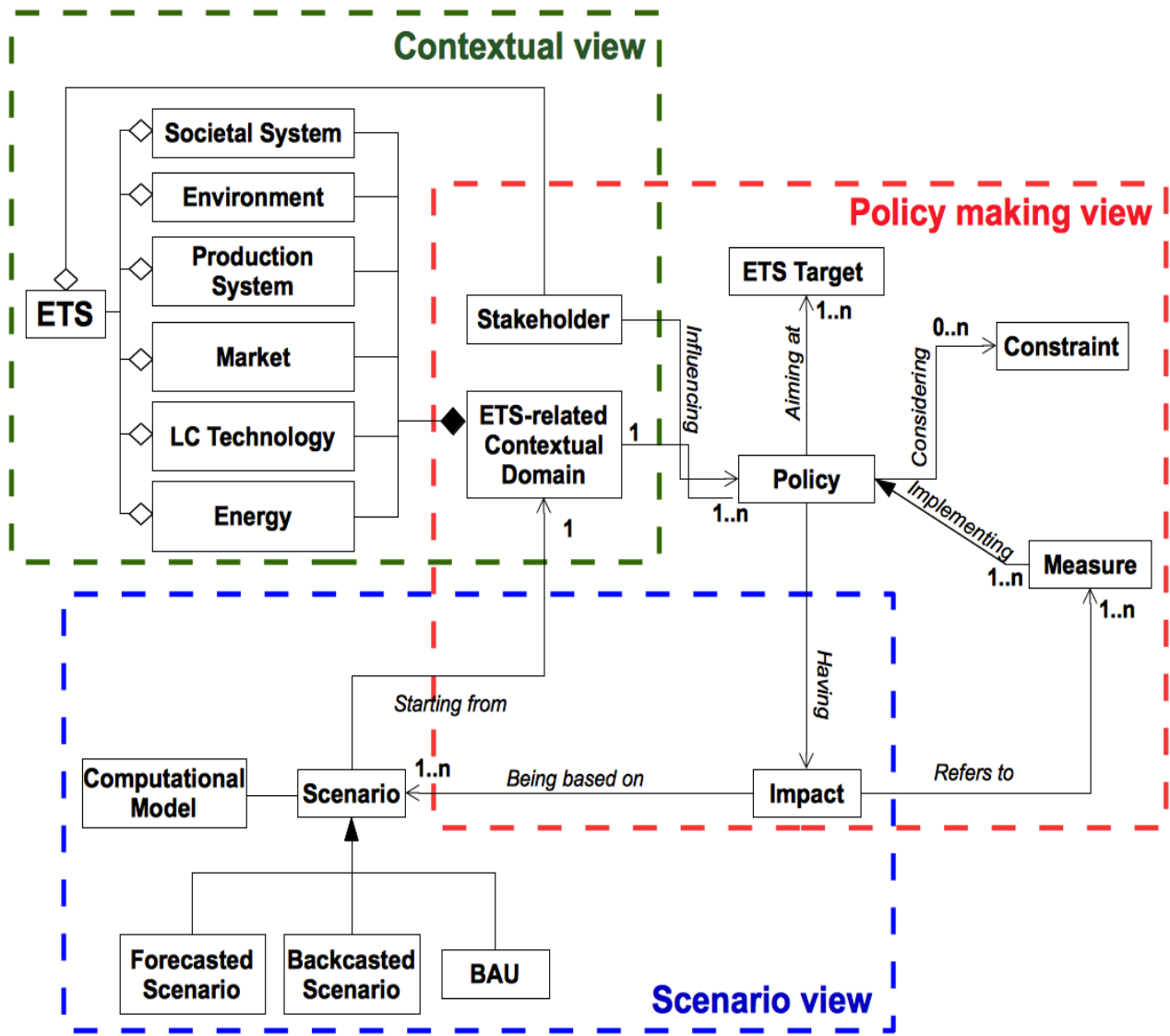

Fig. 5. A representation of foundational ETS-related concepts by using the UML notation. Concepts are organized according to the contextual view, the policy making view, and the scenario view.

The EREON upper level ontology can be compared in its function to other abstract models that have been originally conceived for the business domain, such as HDQM [Batini et al., 2011] and REA [McCarthy, 1982]. However, the primary objective of our model is to represent the ETS-related domain through general concepts and, at the current stage of our work, this does not deal with the business-specific aspects addressed by the cited models (e.g., organization, process, and resource).

Instead, with respect to other existing upper level ontologies, such as SUMO [Niles. Pease, 2001], DOLCE [Gangemi et al., 2002], and Cyc [Reed et al., 2002], the foundational ETS-related concepts share the objective to describe general concepts. However, whereas the above-mentioned upper level ontologies are valid for all the domains as they are intended for ontologies integration, here the focus is on the development of ETS-related ontologies. For this reason, the EREON upper level ontology can be easily derived from, for instance, the SUMO upper level concepts (e.g., abstract entity, physical entity, agent) but it provides general concepts (e.g., ETS Stakeholder) still needing to be further refined (e.g., Energy Company IS_A ETS Stakeholder). Indeed, our purpose has been to define new high-level but domain specific concepts easy to be used and, in case, extended by experts, stakeholders and practitioners of the ETS domain. A step forward to this work could be to provide, for example, a SUMO-compliant version of EREON to increase ontology portability.

A brief description of the three categories of our model follows.

**5.1.1 Contextual View**

The main concepts of this category are the ETS-related contextual domain, the Emissions Trading System, and the stakeholder concepts. The ETS-related contextual domain defines the scope of the overall ontology as it gathers the entities providing the contextual information at the basis of the policy's and of the ETS scenarios development. The Emissions Trading System concept defines the ETS scope through the concepts strictly referring to the EU directives [Grubb, Neuhoff, 2006] [Ellerman, Buchner, 2007] (e.g., gases other than $CO_2$ are outside this specific scope even if they are in the scope of the whole ontology). Finally, the stakeholder represents the actor who has an interest in a policy and can influence its definition. The other foundational concepts of this view are presented in Table 2 of Appendix A.

As a first result for the population of the contextual view of EREON we have encoded the concepts derived by the NIMs [WR3], a data collection mechanism by the European Community regarding, among others, energetic consumptions and GHG emissions from the plants under ETS regulation. Central to the NIMs data structure is indeed the plant concept, formalized as a part-of the Emissions Trading System concept. Indeed, Fig. 6 shows exemplary links between the NIMs and EREON by means of a mapping table. Of course, there are some EREON concepts that are not present in the NIMs (e.g., in case of abstract concepts). For instance, the Emissions concept has been introduced in EREON to better represent the ETS knowledge and to enrich the semantic network behind it, but it is not directly included in the NIMs. Then, a plant is identified through the NACE (Nomenclature statistique des Activités économiques dans la Communauté Européenne) code, a classification scheme for economic and industrial activities adopted by the EU States (last update 2006, NACE Rev. 2, CE law n. 1893/2006 - European Parliament). Also, the ETS activity property refers to Attachment 1 of the ETS guidelines aiming at the identification of the activities that directly contribute to GHG emissions. Both the NACE code and the ETS activity name are fundamental to relate plants data at national and EU levels and to enable various levels of aggregation for specific and/or holistic (cross-EU nation) studies.

The most interesting properties plants owners are required to transmit are about: fuels, emissions, heat, electricity, sub-plants and products. These groups of concepts constitute the seeds for even more specific concepts; this may be due to the various metrics that can be used in the quantification of some parameter, and to the peculiarities of industrial macro-sectors of a given nation due to the technologies or the types of fuels they use. For example, in the Italian steel industry there is a massive usage of electric ovens (37 out of 39 sites analysed), while this might not be the case for other EU countries. Therefore, these technological aspects have to be taken into account in a EU-level industrial analysis.

The contextual part of EREON can be further extended considering, for instance, concepts related to the ETS-market. Here, the existing ontologies to be considered and, in case, adapted are those related to other trading systems, like [Alonso et al., 2005], [Lavbic, Bajec, 2012], [Ying et al., 2013], and [Zhang et al., 2000].

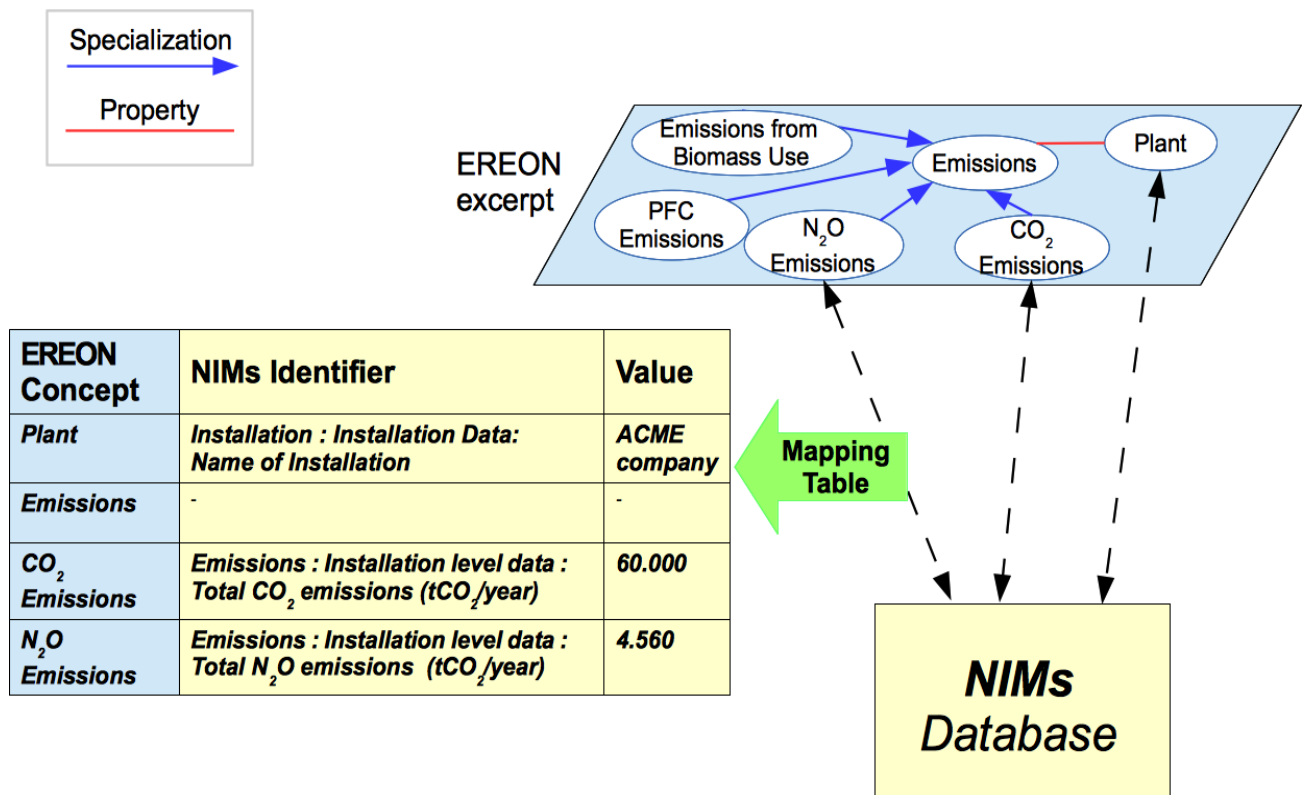

Fig. 6. Exemplary links between the NIMs and an EREON's excerpt by means of a mapping table. The EREON's excerpt shows also a fragment of the semantic network characterized by the plant properties and specialization relationships.

### 5.1.2 Scenario View

This category (Table 3 of Appendix A) is essentially identified by the scenario concept as the root for more detailed concepts related to the simulation-based and/or statistical analysis of ETS. In the energy/environmental market analysis domain, an ETS scenario represents a future projection of the ETS, obtained by means of some computational model. For example the Business As Usual (BAU) concept represents the analysis technique that relies on the assumption that current trends of the status of ETS related entities are maintained, without additional constraints. These projections allow evaluating the impact of some policy on the ETS. The scenario view represents the scope of the studies about impact analysis of political strategies on energy, environment and towards a sustainable economic development. For example, this kind of studies are carried out periodically by ENEA on behalf of the Italian Ministry of the Economic Development and consist of quantitative evaluations of political measures through mathematical models and techniques for scenarios generation.

These studies concern a larger context than ETS, as they analyze the introduction of Best Available Technologies [WR20], including Low Carbon technologies, and account also for non-ETS industries and non-industrial sectors like tourism and the residential one.

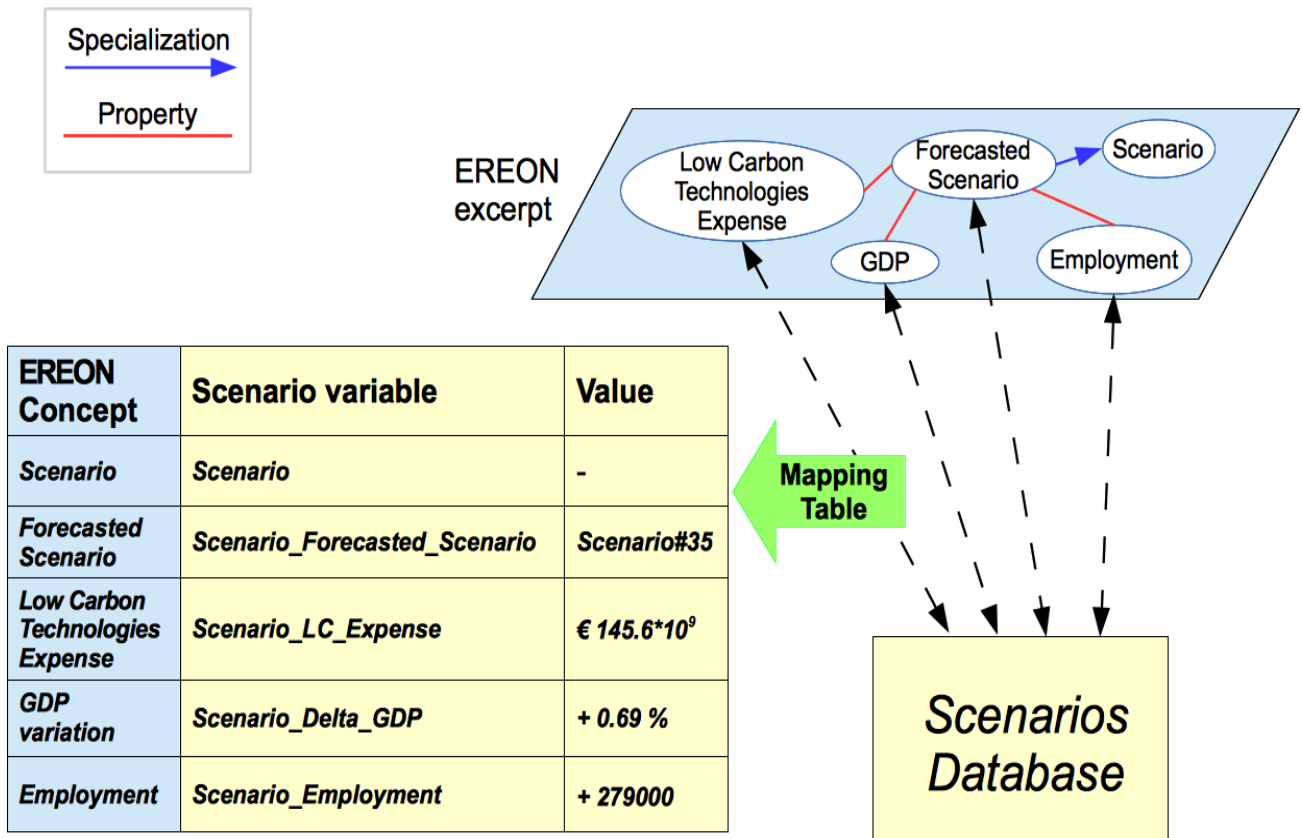

Fig. 7. Exemplary links between a database including scenarios' data and an EREON's excerpt by means of a mapping table.

Our aim is to extend the scenario view to represent knowledge related to these studies in more details. Such ontology will be used to solve interoperability problems occurring, for example, when integrating scenarios data from technological, environmental and economic models, which are implemented by different tools [ENEA, 2013]. Furthermore, the scenarios are not formally linked to the policies originating them in these modeling tools. The EREON ontological foundations can be used to this purpose.

In Fig. 7 we show an example of linkage between a database containing scenarios' data and some concepts of the EREON ontology with root in the scenario view. The scenario, identified by "Scenario#35" in the database, includes data related to the quantification of the impact of low carbon technologies investments on GDP and employment [ENEA, 2009] by the year 2025. In particular, the results given have been computed by assuming a certain total expense by the Italian families, in the years 2007-2020, that accounts for the cost of low carbon electrical appliances and that for the energy consumptions.

**5.1.3 Policy Making View**
This category (Table 4 of Appendix A) includes five main concepts, concerning the social planning aspect of the ETS-related domain of interest. These are: the ETS policy, the ETS target, the constraint, the measure, and the impact concepts. Specifically, an ETS policy is a governmental regulation aimed at achieving one or more desired targets related to reduction of $CO_2$ emissions by means of a set of measures and by considering a set of (national and international) constraints.

The purpose of this view is to represent the various policies related to ETS, for example, through: social (e.g., demographic, infrastructural, residential), economic (e.g., import-export, costs of technologies, costs of governance), and environmental (e.g., effects on climate, energy consumptions, emissions coefficients) indicators. These indicators will be linked to the energy, environmental and economic models used to evaluate them and belonging to the scenarios view.

In Fig. 8 we show an example of linkage between a database containing policies' data and some concepts of the EREON ontology with root in the policy view. In particular, the focus is on the socio-economic evaluation of the impact of the "incentives on renewables" measure within the "directive 2009/28/EC" policy, whose final ETS target is to achieve "21% emissions reduction" [ENEA, 2009]. Such socio-economic impact can be computed through some scenarios including the forecasted scenario named "Scenario#35" (see Fig. 7).

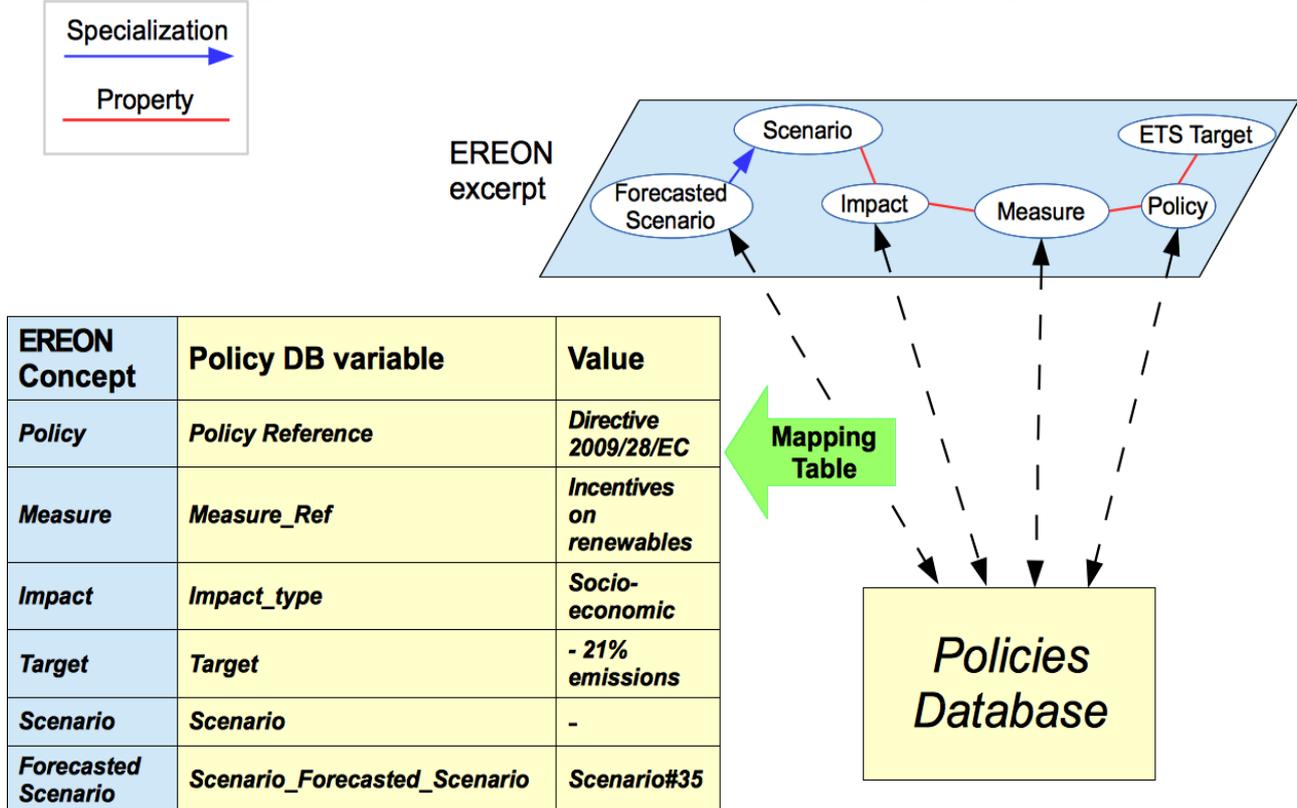

Fig. 8. Exemplary links between a database including information concerning low carbon society policies and an EREON's excerpt by means of a mapping table.

**5.2 EREON Evaluation**

According to [Burton-Jones et al., 2005], an ontology has to be evaluated with respect to four characteristics: *syntactic quality*, *semantic quality*, *pragmatic quality*, and *social quality*.

The Protégé ontology management system allows coding the EREON ontology into correct OWL language [WR21] and, consequently, this ensures the *syntactic quality*.

The *semantic quality* of EREON derives by using the FaCT++ reasoner [WR8] [Tsarkov, Horrocks, 2006] to check the absence of contradictory concepts. This tool is available with Protégé.

The *pragmatic quality* of an ontology concerns fidelity, relevance and completeness. In EREON, fidelity is guaranteed by the domain experts that have contributed to the ontology definition, and by the fact that most part of the ontology is based on NIMs baseline data [WR3] and other references widely accepted by the ETS-related analysts. *Relevance* has been evaluated by successfully implementing semantic services making use of EREON (see next sections). Concerning *completeness*, at the current stage, the EREON structural ontology is focusing mainly on the contextual view, and consists of 161 concepts, 43 object properties, and 27 data properties. It is provided as an OWL project downloadable from [WR21].

Finally, *social quality* concerns how much the ontology is accepted by the reference community. At the moment such community is composed by the domain experts that contributed to its development. However our intention is to make it open to contributions from other experts in the area, for example, by enabling them to directly access to it through the Internet (e.g., by means of web Protégé [WR22]).

**6. ETS Semantic Services**

Based on the plant instances, data studies at industrial macro-sector level are possible, focusing, for example, on energy consumptions, production, adopted technologies, and related gas emissions. However, preconditions for such analyses are the correctness and consistency of the base data. On the other hand, the analyses that can be performed may use recurring low levels queries, such as threshold-based data classification, or impact analysis on ETS-related data of some new regulation.

The formalization through semantic rules of the correctness/validation criteria for the data, and of the recurrent or most interesting low-level queries, can be useful to speed up the data analysis process and be a basis for more advanced analysis services. To this aim, we have built a catalogue of rules that have originated from the requirements of the Italian Ministry of Economic Development [WR2] in the context of our research project.

Focusing on the NIMs data structure, currently we have formalized 32 rules that we have organized in a taxonomy, based on their function, and whose description and examples will be presented in the sequel. The taxonomy is presented in graphical form in Fig. 9 and originates from an elicitation process involving both domain experts, providing their knowledge about the ETS, and knowledge engineers, providing their expertise concerning SWRL and logic programming. Following [Murthy, 1998] [Quinlan, 1987] [Quinlan, 1993], we have modelled a set of questions and answers as a decision tree that, finally, map to sets of rules. For instance, the rule category *ETS Data Correctness* originates from the following decision tree path: ***Q**. What is your objective? **A**. ETS data acquisition. **Q**. What do the data need? **A**. Verification and validation. **Q**. What should be checked? **A**. Correctness.*

Both questions and answers, from which the semantics of the rules classes arise, reflect the users analysis activities and needs. Indeed, the aim of the taxonomy is to provide users with a guideline for searching and creating rules.

It should be noted that, as the ontology, the taxonomy of the rules could evolve in time to gather changes in the ETS domain and new requirements from stakeholders and domain experts.

The taxonomy in Fig. 9 constitutes also the ground for semantic services definition. Indeed, the two services we developed handle the categories of rules at the second level of the hierarchy. As the taxonomy is extendible at all levels by means of the question-answering approach mentioned above, new services could be built after new rules are defined, whenever the need for them arises. In general, the role of a semantic service refers to the rules type that the service handles, hence to the activity where these rules are used. Indeed, the rules implement the main part of the logic of a service and identify its responsibility. However, a new service should be introduced whenever its logic, other than the rules, is not implemented by the existing services. In this case, once the rules have been formalized, the rule engine can run them without further programming effort.

A detailed description of the rules mentioned above is given in the following subsections. The rules are presented in the SWRL-like syntax [WR12] and in a simplified form to ease human readability. In particular, the definition of some unary predicates has been omitted whereas their meaning is clear from the context and ternary predicates have been used to present the rules in a more compact way.

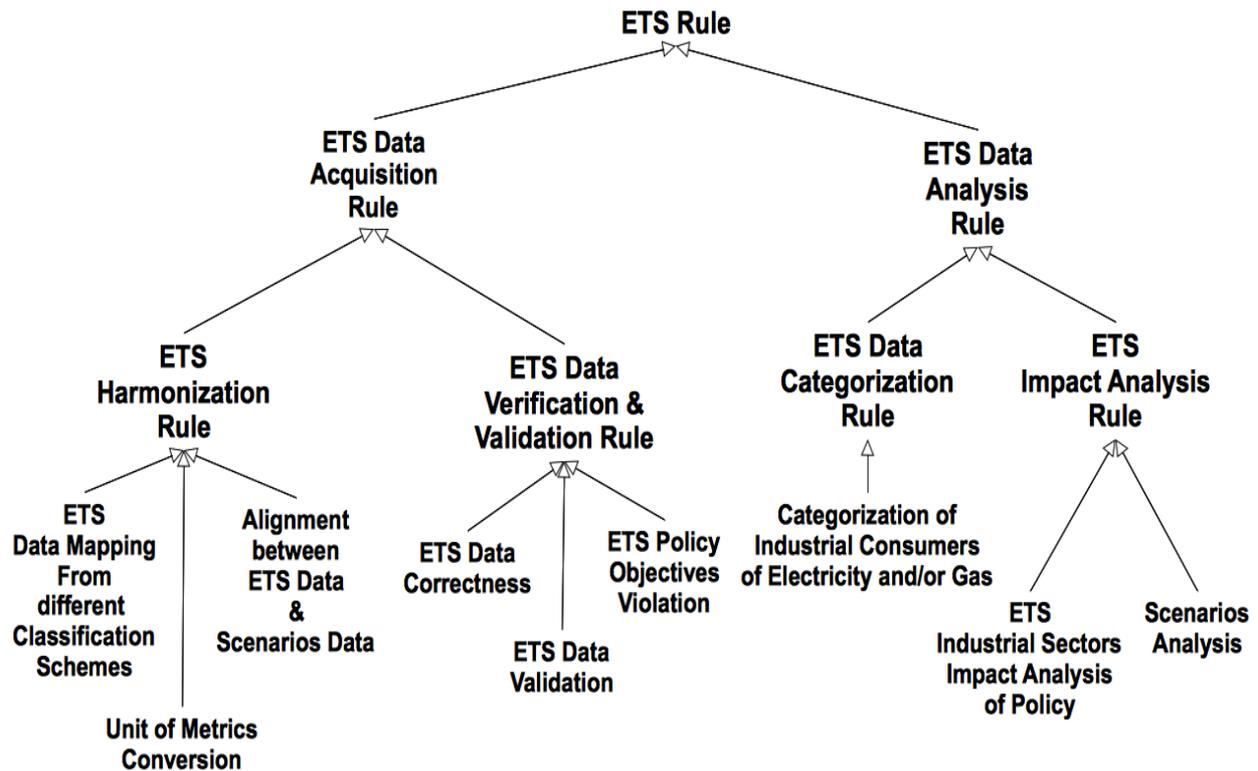

Fig. 9 ETS Rules Taxonomy.

**6.1 Data Acquisition Service**
*ETS Data harmonization rules*
Rules belonging to this category allow relating data coming from different sources (or models) to the aim of verification/validation and/or aggregation operations. In the ETS context, we have identified the following two different rule types belonging to this category.
- ***ETS data mapping from different classification schemes.*** Different classification schemes exist, for example, for activities and products of the plants. The EU-ETS regulation establishes that activities have to be classified through the NACE code. However, benchmark values (in terms of $CO_2$ allowances) have been defined per products of those activities [WR23], in order to ensure that the free allocation of emissions allowances takes place in a manner that provides incentives for reductions in greenhouse gas emissions and energy efficient techniques. Therefore, means to relate industrial products data with activities data are necessary. One option is to use the PRODCOM classification, defined through a CEE regulation (n. 3924/91). This consists of a harmonized list of the industrial products to be monitored by the EU member states (through their statistics observatories). Generally, PRODCOM codes, described by 8 digits, have been constructed starting from NACE codes, which consist of 4 digits. However, from an accurate study within the project, it resulted that a NACE code may originate none, one or more PRODCOM codes, corresponding to benchmarked products; conversely, a PRODCOM code may derive from none or one NACE code. Specific correspondences between these codes have been identified and encoded though semantic rules.

For the sake of conciseness, we present here a partial but meaningful formalization of this rule:

*Every* product identified by a PRODCOM code **is** originated from an activity identified by a NACE code.

$product(?x) \land prodcom(?y) \land hasProdcom(?x,?y)$

$\quad \rightarrow derivesFromNace(?y,?z) \land Nace(?z) \land activity(?t) \land hasNace(?t,?z)$

- **Units of metrics conversions**. These rules allow comparing data values expressed in different metrics in order to check their conformance with national or international standards and also enable aggregation-based analysis. These rules are especially useful to harmonize emissions values originated by different fuels, in order to compute an overall analysis at plant or industrial sector levels. For the Italian system, the metrics and their conversion factors related to heat, electricity, and fuels are established within the National Energy Balance by the Ministry of Economic Development. We have defined rules based on these units of metrics conversions, like that 1 ton of coal (tec) corresponds to 0.7 tons of petroleum (tep) and that 1 tep is equal to 41.868 GJ in energy. The formalization of a rule based on the first conversion above is presented in the following.

*h tons of coal (tec)* **corresponds to** *0.7\*h tons of petroleum (tep).*

$(coal(?x) \land coalQuantity(?x,?y) \land hasValue(?y, h, tec) \land$
$petroleum(?w) \land petroleumQuantity(?w,?y))$

$\quad \rightarrow hasValue(?y, 0.7 * h, tep))$

- **Alignment between the ETS data structure and scenarios data structures**. These types of rules are conceived to support projections and scenario-based analysis activities finalized to quantifying the impact of political measures in the long run. To build scenarios, data about past and present trends are required, as the aim is to evaluate deviations from the *Business As Usual* (BAU)-type studies. Scenarios are constructed by using ad hoc software tools (e.g., TIMES [WR4] and GAINS [WR5]), for example, to simulate the competitiveness of fuels or technologies in the energy market and also the impact of environmental policies, such as the introduction of a tax on the emissions. Therefore, input and output data structures from and towards these tools have to be aligned with the ETS data structure. Rules that allow such an alignment are tool-dependent and usually are either of data mapping or aggregation types. Currently, we have encoded rules to support alignment with tools dealing with technological scenarios (e.g., TIMES) and with ecological scenarios (e.g., GAINS). We present here one example of such an alignment whose aim is to guarantee that scenarios are produced from the same base year and the same descriptive data, by all the above tools.

**For a** *given year,* **for a** *given ETS subsector, the sum of energy consumptions by the companies must be equal to the overall energy consumption of that ETS subsector in that year.*

$(etsSubSector(?x) \land year(?y) \land etsSubSectorEnCons(?x,?y,?e) \land$
$etsSubSectorComp(?x,?z_1) \land .. \land etsSubSectorComp(?x,?z_n) \land$
$etsSubSectorCompanyEnCons(?z_1,?y,?e_1) \land ..$
$\land etsSubSectorCompanyEnCons(?z_n,?y,?e_n)) \rightarrow$

$$equal(?sum(?e_i), ?e)$$

*ETS data verification and validation rules*
Rules belonging to this category are "control" rules, either to support data quality verification or the validation of an ETS regulation or of a policy, or else to detect violations of some desiderata.

- **ETS data correctness**. These types of rules aim at ensuring usability and reliability of the input data for analysis purposes. They could be derived from the knowledge of the domain.
  We have defined a set of rules to verify the reliability of the gas emissions data by a certain plant, based on the fuel types used by that plant.
  Specifically, these rules relate the value of the emissions of some gas, produced by a certain fuel, with the maximum emission factor that can be produced by that fuel. Maximum values are generally established by official national or international standards. Indeed, for $CO_2$ emissions, these reference values are being computed periodically by the United Nations Framework Convention on Climate Change (UNFCCC). Our rules are based on a table with evaluations of $CO_2$ emission factors per fuel type, currently available from the Italian Ministry of Environment web site [WR2], to hold for the Italian context from January 2011 to December 2012.
  Here we provide one example of such rules:

  **The** $CO_2$ emissions value originated by 1 TJ of natural gas burning **must be** less than or equal to 56 $tCO_2$.

  $(CO_2Emission(?x) \land naturalGas(?z) \land naturalGasQuantity(?z,?y) \land$
  $hasValue(?y, 1, TJ) \land originatedBy(?x,?z) \land hasValue(?x, k, tCO_2))$
  $\quad \rightarrow lessThanOrEqual(?k, 56)$

- **ETS data validation**. These types of rules may be derived by the ETS regulations. Examples include rules to check that the $CO_2$ emissions transmitted by a plant are within its ETS allowance and rules to check that these data have been certified by a third party authority.
  We present here the formalization of the first constraint:

  **For every** year, **for every** plant, the verified $CO_2$ emissions value **must be** less than or equal to its allocated $CO_2$ emissions value.

  $(year(?x) \land plant\ (?y) \land CO_2Emission(?w) \land CO_2Emission(?z) \land$
  $hasVerifiedEmission(?x,?y,?w) \land hasAllocatedEmission(?x,?y,?z))$
  $\quad \rightarrow lessThanOrEqual(?w,?z)$

- **ETS policy objectives violation**. Rules of this type may encode expectations from some policy or industrial strategy. These expectations may be by various stakeholders at different granularity: those by politicians at national or EU level, who reason about macro-effects in the long run of their policies, and those by plants owners who need to negotiate $CO_2$ allowances in the ETS. Examples from plants owners include rules codifying a dependency between the number of allowances acquired by a plant and its gas emissions production.
  In both cases, formalizing the objectives of some policy/strategy through rules may be useful to highlight violations or failures over time.

For these types of rules, we have defined templates, as concrete rules may be instantiated after specific studies. Here we provide an example related to a political measure.

**If** the $CO_2$ price is increased by V1 percent in the year Y1 with respect to the year Y0, **then** the $CO_2$ emissions of sector W should decrease of V2 percent in the year Y2.

$(year(?y_0) \wedge year(?y_1) \wedge year(?y_2) \wedge lessThan(?y_0,?y_1) \wedge lessThan(?y_1,?y_2) \wedge annualCO_2price(?x_0,?y_0) \wedge annualCO_2price(?x_1,?y_1) \wedge annualCO_2price(?x_2,?y_2) \wedge equal(?x_1,?x_0 + v1 *?x_0) \wedge Sector(?w) \wedge hasCO_2Emission(?c_1,?w,?y_1) \wedge hasCO_2Emission(?c_2,?w,?y_2)) \rightarrow equal(?c_2,?c_1 - v_2 *?c_1)$

**6.2 Data Analysis Service**

*ETS data categorization rules*
Rules belonging to this category are queries and aim at aggregating and classifying raw data.

- ***Categorization of industrial consumers of electricity and/or gas.*** These types of queries support in the identification of the industrial sector or consumers that can be affected by a change to some variable, such as electricity/gas price.
  Often, the final electricity price includes additional charges to benefit various public targets, such as the administration costs regarding TSO (Transmission System Operator) and regulatory authority, the financial green certificates or the subsidies to RES (Renewables Energies), energy efficiency. However, these additional costs are not the same for all consumers. The Italian Regulatory Authority for Electricity and Gas (AEEG) identifies four non-domestic electricity consumer groups on the basis of consumption thresholds, and these groups have different charge policies. They are:
  - consumers with a monthly consumption within 4 GWh;
  - consumers with a monthly consumption from 4 GWh to 8 GWh;
  - consumers with a monthly consumption from 8 GWh to 12 GWh;
  - consumers with a monthly consumption greater than 12 GWh.
  
  Categorization criteria from the various EU countries can be correlated through the ontology for a holistic analysis.
  Here we provide the formalization of the rule concerning the additional charge to enable decommission of nuclear plants (named A2) in the final Italian electricity price determination.

  **Every** consumer with a monthly consumption greater than 8 GWh and less than or equal to 12 GWh **has** lower taxation with respect to component A2 than a consumer with a monthly consumption less than or equal to 4 GWh **and has** higher A2 taxation than a consumer with a monthly consumption greater than 12 GWh.

  $(consumer(?c_1) \wedge hasMonthlyConsumption(?c_1,?m_1, GWh) \wedge\\ greaterThan(?m_1, 8) \wedge lessThanOrEqual(?m_1, 12) \wedge\\ consumer(?c_2) \wedge hasMonthlyConsumption(?c_2,?m_2, GWh) \wedge\\ lessThanOrEqual(?m_2, 4) \wedge consumer(?c_3) \wedge\\ hasMonthlyConsumption(?c_3,?m_3, GWh) \wedge greaterThan(?m_3, 12) \wedge\\ hasA_2Taxation(?c_1,?t_1) \wedge hasA_2Taxation(?c_2,?t_2) \wedge hasA_2Taxation(?c_3,?t_3))\\ \rightarrow greaterThan(?t_2,?t_1) \wedge greaterThan(?t_1,?t_3)$

*ETS impact analysis*
Rules of this category are queries finalized at a *what-if* analysis.

- **ETS industrial sectors impact analysis of policies**. These types of queries are concerned with changes in gas/electricity prices (or other fuel price) or a carbon tax and their impact on the ETS industrial sectors. This generally requires answering to lower level queries.
  For example, queries based on the NIMs data may be:
  - quantifying the number of firms with emissions up to a certain threshold;
  - identifying the industrial sectors with $CO_2$ emissions up to a certain threshold;
  - identifying the industrial sector with energy consumption up to a certain threshold (in this case, it will be possible to study the impact of rate component of electricity or gas prices).

  Here we provide the formalization of the rule concerning the emissions by a given industrial sector.

  ***Every*** *industrial sector* ***has*** *an overall $CO_2$ emissions less than or equal to X.*

  $$(sector(?w) \land CO_2Emission(?x) \land hasEmission(?x,?w))$$
  $$\rightarrow lessThanOrEqual(?x, X)$$

- **Scenarios analysis**. The long run evaluation of policies may be conducted through scenarios. These describe hypothetical processes or sequences of events that could develop over a period of time, originating from some conjecture. Indeed, scenarios integrate radical departures from trends, due, for example, to technological breakthroughs, or major shifts in human behaviour or to political investments. Concerning the ETS context, we have identified a group of rules for querying scenarios that are more or less abstract to be applied to various studies.

  For example, queries based on the comparison of scenarios:
  - identifying the impact of changing low carbon technology;
  - identifying the subsector or subsectors impacted by the variation of energy prices (natural gas, electricity, etc.);
  - identifying the main technologies used in the sub-sectorial industries;
  - identifying the subsector or subsectors (negatively) impacted (e.g., decrease of employment rate) by the variation of energy prices (natural gas, electricity, etc.).

The last example is formalized below.

***For each*** *industrial sector,* ***if*** *the energy price has increased its value after five years* ***then*** *the employment rate has decreased in the same period of time.*

$$(sector(?w) \land year(?y_0) \land year(?y_1) \land equalTo(?y_1 - ?y_0, 5) \land$$
$$annualEnergyPrice(?x_0,?y_0) \land annualEnergyPrice(?x_1,?y_1) \land lessThan(?x_0,?x_1)$$
$$\land annualEmployment(?w,?y_0,?e_0) \land annualEmployment(?w,?y_1,?e_1))$$
$$\rightarrow lessThan(?e_1,?e_0)$$

## 7. Rules Evaluation

The EREON ontology and the semantic rules library are the specification for the semantic services described in Section 6. One technical precondition for the adoption of such a specification is the correctness of the rules and the usability assessment of semantic technologies in the ETS domain. To this aim, in this section we present the software environment we used for rules evaluation and discuss some results we obtained in this respect.

**7.1 Tools Support**

To the purpose of this experimentation, we have codified the rules in SPARQL [WR13]. In general, we suggest to keep the codified rules separate from the ontology in order to provide more flexibility to the various users in their choice of the specific set of rules they might be interested in (in principle, not all users share the same set of rules).

We developed a Java-based software environment integrating ESS with a relational database reflecting the NIMs schema [WR3]. The software was built according to the architecture described in Fig. 4. Specifically, the components of the architecture were developed as follows:

- *Custom application*: Java component responsible for generation of the NIMs data and execution of the ESS rules specified in SPARQL [WR13]. For privacy protection, to the purpose of this experimentation, we decided to use simulated NIMs data rather than the real ones. Therefore, an automatic data generator was implemented to generate them. Such generator uses the java libraries allowing the generation of random numbers or following statistical distributions.
  This component also integrates a SPARQL client for the rules execution.
- *Mapping service*: component that allows accessing the NIMs data stored in the DB, in order to fire the SPARQL rules. This component was realized through the D2RQ technology [WR17], based on the Jena [WR24] engine, and is also freely available with Java APIs. The EREON to DB schema mapping was quite easy to carry out as the NIMs structure nicely fits with the contextual view part of EREON.
- *Relational DB*: DB containing the NIMs data, realized by PostgreSQL [WR25].

It should be noted that, in our framework, the separation between the methodological approach and the technical aspects allows a seamless migration from a technical solution to another one.

**7.2 Empirical Study**

The *object* of this empirical study are the proposed ESS rules with the *purpose* of evaluating their effectiveness and the performance in their execution. The study has two *perspectives*: (i) the *researcher*, who wants to demonstrate the usefulness and applicability of semantic tools in this new application field, and the (ii) *ETS analyst*, who wants to evaluate the support given by these tools in his/her activities. This section summarizes the results of such an empirical study. Data concerning the empirical study can be downloaded from [WR26].

*Context description*

A representative subset of four rules has been selected from the library for evaluation. Specifically, these rules are reported in the Appendix B with SPARQL formalization. The choice of the rules has been made by considering both the main types of our rules classification (i.e., *ETS data acquisition* and *ETS data analysis rules*) and their applicability to NIMs data. The datasets over which to perform the queries were generated according to two dimensions: the number of plants, to simulate temporal evolution/updates of ETS data, and the number of rules violations to obtain different results.

The generated data values belong to the database tables that are impacted by the four selected rules.

The ranges of such values were predetermined by domain experts and according to the following criteria:
- even if simulated by random, Poissonian and Gaussian distribution, data samples should be likely in real world scenarios;

- number of rule violations should be in accordance with the specific empirical test.

*Objectives*
The study intends to provide answers to the following research questions:
RQ1.   How do semantic queries perform over ETS data?
RQ2.   How does their performance scale after temporal updates of the ETS data?
RQ3.   Does the performance of semantic queries relate to their result (success or violation)?

*Study settings*
Both the software application and the database were installed on an Intel Core i7 2.8 GHz machine, with 6 Gb of RAM and Window 7 operating system, JDK 1.7, D2RQ 0.8.1 [WR17], and PostgreSQL 9.1 [WR25]. Response times for each rule execution were measured using the system time utility, i.e., getting the user CPU time. To avoid bias introduced by randomness of the simulated NIMs data, we replicated the generation of the datasets 10 times.

*Study results*
This subsection reports results of the empirical study, aimed at answering the research questions stated beforehand.

To answer to RQ1 we generated 10 datasets with data from 3000 plants, as this is currently the average number of Italian companies subject to ETS regulations. Fig. 10 shows boxplots synthetizing the performances of the four chosen queries on the given datasets. No significant differences in the results were observed for the various datasets and in all cases the resulting performance is acceptable (the maximum time to obtain an answer is of about 6 minutes).

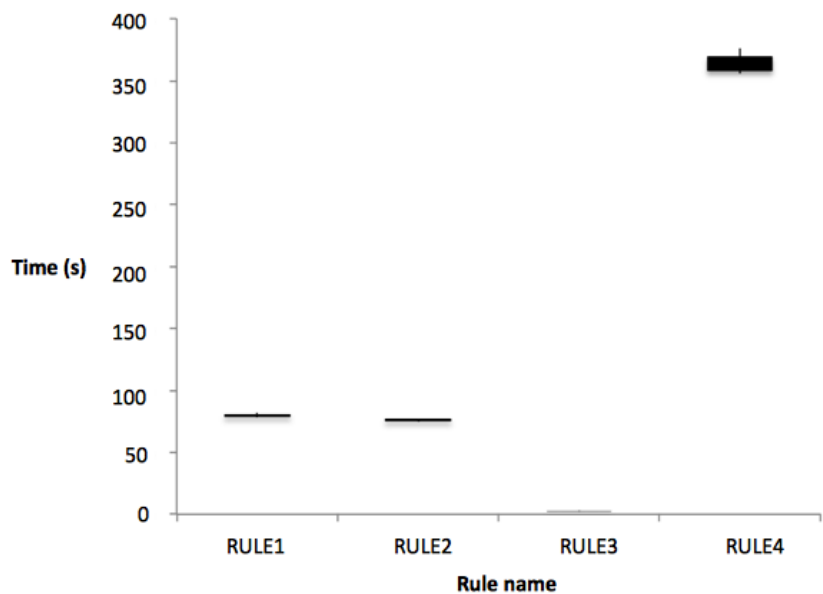

Fig. 10: Performance of rules verification on an average size dataset. Standard Deviations (SD)s: SD(RULE1) = 1.1; SD(RULE2) = 0.8; SD(RULE3) = 0.3; SD(RULE4) = 7.0.

To answer to RQ2 we generated 10 datasets for each of the following sizes: 10, 100, 500, 1000, 2000, 3000, 4000, 5000 and 20000 plants. This choice is motivated by the fact that the number of plants is the most sensitive variable to variation over time. Fig. 11 shows the performances obtained for each rule and each size where mean values over the 10 datasets of that size were considered. One can recognize that significant time variation may be present (see Rule 1 and Rule 2). Indeed, for Rule 3 and Rule 4 the performance is independent on the size of the dataset as these rules are not directly concerned with plants data. The highest value obtained for the performances in all runs was of about 10 minutes on 20000 plants, and this is still affordable as rules verification is an occasional activity.

To answer to RQ3 we generated datasets to test Rule 1 and Rule 2, where the number of plants was fixed to 3000 and introduced a number of violations equals to: 0, 10, 100, 1000, 2000 and 3000. As expected, we obtained the same performances for all datasets.

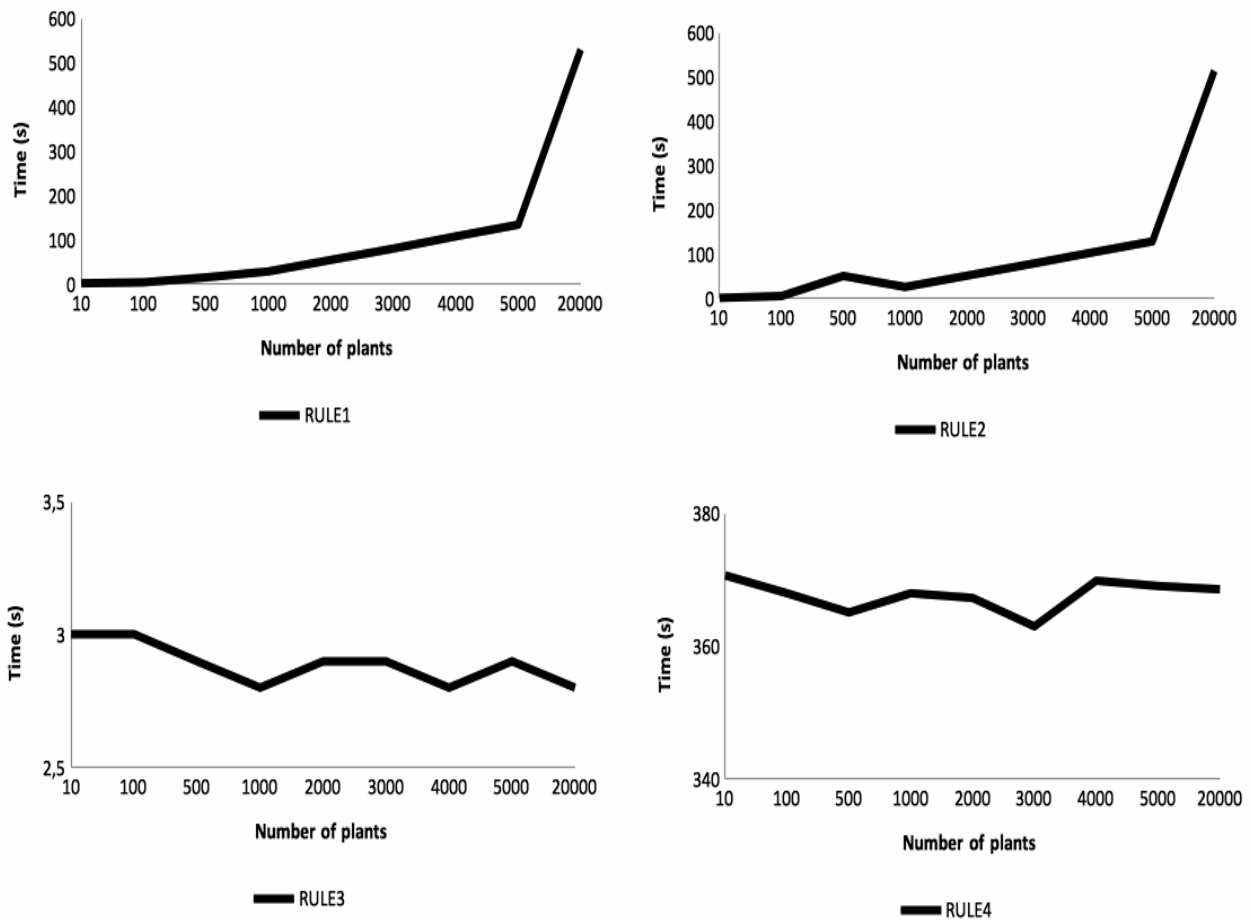

Fig. 11: Performance of rules verification on datasets of increasing sizes.

*Threats to validity*

Threats to *construct validity* deal with limitations on the experimental set up. In our case we have dealt with a subset of EREON concepts and of the rules to process their instances. Indeed, for this we focused on around 8 entities of EREON and their properties, such as: *plant*, *industrial sector*, $CO_2$ *emissions*, *ETS scenario*, $CO_2$ *price*, *employment*, *natural gas*, and *energy price*. However, it can be noticed that in this subset we included the most frequent entities involved in the rules presented in Section 6 and the one most sensitive to evolution and update operations (the *plant*), enough to address the objectives of our experimental analysis.

Further threats to *construct validity* are related to the number of plants we have considered, which may impact on the performance results in time but also on disk size. Since our case study refers to the Italian ETS, which collects NIMs data from about 3000 plants and, currently, the EU-ETS [Stephan et al., 2014] covers "only" 11000 installations, we have considered the boundary number of 20000 installations. According to our experimentation, this number leads to an amount of records perfectly manageable by a relational database and by the semantics technology we used on top of it. Furthermore, by increasing the number of plants up to 50000 (as we actually did), which is much higher than any realistic case in Italy and Europe, we obtained the time performances for each rule as in Table 1.

| **Rule** | **Time performance** |
|---|---|
| Rule 1 | 2075 sec |
| Rule 2 | 1980 sec |
| Rule 3 | 4 sec |
| Rule 4 | 497 sec |

Table 1. Time performance on rules verification for a dataset covering 50000 installations.

Even in the case of a huge increase of the ETS data, semantics-based technologies might still be usable, according to some benchmark results such as those provided by the Ontop technology [WR18], and works discussing SPARQL over distributed databases [Zhou, Zheng, 2011], [Liu et al. 2014].

*Internal validity* threats, instead, may deal with the randomness of the results and with the performances of the hardware/software we used. Such a threat was limited in two ways: (i) deploying the whole application on a single dedicated machine, not used for other purposes; (ii) replicating each run 10 times and perform analyses over results from all runs.

Finally, *external validity* threats concern with generalizability of the results. These have been addressed by considering realistic values in the generation of the input datasets, as suggested by domain experts. Indeed, real data had to be protected for privacy reasons.

## 8. Conclusions

Management of knowledge related to the low carbon society is a complex and open issue for the research community. This is due to the multi-disciplinary nature of this evolving domain encompassing technological and economical areas from energy to environment, from industrial technologies to social systems.

Semantic technologies can provide a significant support by means of a formal and integrated environmental modelling approach. Open issues in such direction concern the lack of available semantic specifications and real applicability of semantic technologies due to scalability problems. Here we face such challenges by providing a knowledge base concerning the ETS domain, consisting of the EREON ontology and the ESS rules.

Then we demonstrate the applicability of the semantic approach by means of the implementation of two semantic services: the *data acquisition service* and the *data analysis service*. The ETS knowledge base has been conceived to become an open project. Furthermore, although it has been built for the specific ETS application, the EREON ontology, the ESS rules and the semantics-based architecture can be easily extended to cover a broader domain by considering other LCS concepts.

Indeed, the semantics-based framework contributes to solving the five issues we stated in the introduction that arise from the LCS policies definition and analysis problem. Namely, the EREON ontology and the ESS rules are the result of formal representation and integration of knowledge from the various fields covered by the EU-ETS related domain (1), and have been built with the support of domain experts (2). Also, by their own nature, they concern both structural and behavioural knowledge (3), which is mainly covered by the data from the Italian plants committed to ETS that have been collected through the NIMs (4). Such data are the basis for scenarios-based impact analysis from technological, environmental, societal and economic perspectives of political measures towards ETS objectives (5).

One issue, here, concerns the evolution of the knowledge base. For both ontology and rules, this requires the collaboration of domain experts with knowledge engineers, who are technically skilled people not easy to be overcome. Concerning the specification of new rules, in particular, we deem that when defining a domain-specific technological framework for editing and running queries, flexibility and usability should be balanced. In our case, improving usability, for example through a domain-specific rules editor, would probably result in limiting expressiveness and extensibility. As the EU-ETS domain is rapidly changing, the flexibility of the software to reflect these changes is an important requirement. Therefore, a general-purpose semantics-based technology to support a domain-specific modelling methodology is a good compromise.

Finally, as future work, we intend to address the LCS community to involve more researchers, stakeholders and practitioners in this research work.

**Acknowledgements**

This work has been inspired by an ongoing project activity in favour of the Italian Ministry of Economic Development (MiSE). Many enlightening discussions on these topics with Antonio Bartoloni (MiSE) are kindly acknowledged. Vittorio Rosato is kindly acknowledged for a critical reading of the manuscript and for useful suggestions. Finally, we wish to acknowledge several ENEA colleagues who contributed to improving our work with their constructive comments: Vincenzo Artale, Bruno Baldissarra, Natale Massimo Caminiti, Umberto Ciorba, Ilaria D'Elia, Maria Gaeta, Stefano La Malfa, Sergio La Motta, Carlo Manna, Roberto Morabito, Tiziano Pignatelli, Marco Rao, Pasquale Spezzano, Marco Stefanoni, Giovanni Vialetto, Maria Velardi, and Maria Rosa Virdis.

This work was conducted using the Protégé resource, which is supported by grant GM10331601 from the National Institute of General Medical Sciences of the United States National Institutes of Health.

**Web References**

**APPENDIX A**

| Concept's name | Concept's description |
|---|---|
| ETS-related Contextual Domain | The ETS-related Contextual Domain (ECD) gathers the entities providing the contextual information at the basis of policy's definition and of ETS scenarios development. This domain is inherently multi-disciplinary since related to energy, low-carbon technologies, market, production system, environment, and societal system domains. |
| Emissions Trading System (ETS) | ETS is a scheme for trading greenhouse gas emissions allowance within the EU Community (Community scheme) in order to promote reductions of greenhouse gas emissions in a cost-effective and economically efficient manner [Kumazawa et al., 2009]. |
| Stakeholder | A person, group or organization that has an interest or a concern in a policy. |
| Societal System | Set of social relations that is designed and analysed as a complex of elements characterized by interdependent laws to be respected within a surrounding environment [Parsons, 1949] [Parsons, 1951]. |
| Environment | Set of all elements that can influence directly or indirectly the natural evolution of the human, animal and plant life. The GHG emissions and their concentration in the atmosphere have been increasing since the post industrial revolution, with severe impact on the natural circle. |
| Production System | Firms, companies and enterprises generating products using raw materials, machineries, work, and capital to the aim of selling them and obtain profits. |
| Market | A virtual or physical place or else a process by which the prices of goods or/and services are established. |
| Low Carbon Technology | Technology characterised by low carbon emissions and that should replace high emission machines and utilities. Normally, a low carbon technology is an innovative technology or process. |
| Energy | A property of our physical world on which modern life relies. Energy consumption heavily contributes to $CO_2$ emissions. |

Table 2. Contextual view foundational concepts.

| Concept's name | Concept's description |
|---|---|
| Scenario | A scenario is a coherent, internally consistent and plausible description of a possible future state of the world. It is not a forecast; rather, each scenario is one alternative image of how the future can unfold [WR27]. |
| Business As Usual (BAU) | A scenario based on the assumption that current trends of the status of ETS related entities are maintained, without additional constraints. |
| Backcasted Scenario | A scenario based on the following question: "if we want to attain a certain goal, what actions must be taken to get there?" [Tinker, 1996]. In other words, this simulation starts to describe the desired future and then identifies the direction to be followed connecting the future to the present. |
| Forecasted Scenario | A scenario obtained through forecasting, i.e., the process of predicting the future based on current analysis, technologies and also (implemented, adopted and planned) policies. |
| Computational Model | A technical-economic model of a system. One example for the energy system is the Markal-Times computational model. This model calculates the optimal mix of technologies and commodities and it is enable to evaluate energy plans, environmental policies, climate mitigation scenarios and new technologies in trade-off modes. |

Table 3. Scenario view foundational concepts.

| Concept's name | Concept's description |
|---|---|
| ETS policy | An ETS policy is a governmental regulation aimed at achieving one or more desired targets related to reduction of $CO_2$ emissions by means of a set of measures and by considering a set of (national and international) constraints. |
| ETS target | The ETS target is a desired state of the affairs related to ETS. It drives the ETS policy definition process. |
| Constraint | Any kind of limitation or restriction to be considered in defining an ETS policy. |
| Measure | An action or procedure intended as a means to an end [WR28]. |
| Impact | The estimated effects of policy. |

Table 4. Policy making view foundational concepts.

**APPENDIX B**

Rules selected for the evaluation:

*RULE1* = SELECT DISTINCT ?plant ?fuel_quantity ?ng_quantity ?year WHERE {
            ?x vocab:total_plant_co2_emissions_natural_gas_burning_ref ?nb ;
              vocab:total_plant_co2_emissions_unit_of_measure "tCO2" ;
             " vocab:total_plant_co2_emissions_plant_ref ?plant ;
             " vocab:total_plant_co2_emissions_ng_quantity ?ng_quantity ;
             " vocab:total_plant_co2_emissions_year ?year .
            ?nb a vocab:natural_gas_burning ;
              vocab:natural_gas_burning_quantity ?fuel_quantity ;
              vocab:natural_gas_burning_unit_of_measure "Sm3" ;
              vocab:natural_gas_burning_plant_ref ?plant ;
              vocab:natural_gas_burning_year ?year .
            ?plant a vocab:plant .
     FILTER  (   (2000 * ?ng_quantity) < (1.961 * ?fuel_quantity) ) }

*RULE2* = SELECT DISTINCT ?plant ?aCO2 ?vCO2 ?year WHERE {
         ?x vocab:allocated_co2_emissions_quantity ?aCO2 ;
            vocab:allocated_co2_emissions_plant_ref ?plant ;
            vocab:allocated_co2_emissions_year ?year .
         ?y vocab:total_plant_co2_emissions_quantity ?vCO2 ;
            vocab:total_plant_co2_emissions_plant_ref ?plant ;
            vocab:total_plant_co2_emissions_year ?year .
         ?plant a vocab:plant .
      FILTER  (  ?vCO2 > ?aCO2) }

*RULE3* = SELECT DISTINCT ?sector ?year2  WHERE {
         ?x vocab:co2_price_value ?c1 ;
            vocab:co2_price_year ?year1 .
         ?z vocab:co2_price_value ?c2 ;
            vocab:co2_price_year ?year2 .
         ?y vocab:co2_emissions_by_industrial_sector_quantity ?e1 ;
            vocab:co2_emissions_by_industrial_sector_year ?year1 ;
            vocab:co2_emissions_by_industrial_sector_industrial_sector_ref ?sector .
         ?w vocab:co2_emissions_by_industrial_sector_quantity ?e2 ;
            vocab:co2_emissions_by_industrial_sector_year ?year2 ;
            vocab:co2_emissions_by_industrial_sector_industrial_sector_ref ?sector .
         ?sector a vocab:industrial_sector .

       FILTER  ( ((?c2 - ?c1)/?c1 > 0.1) && ((?e1 - ?e2)/?e1 > 0.001) && (?year2 = ?year1 + 1) ) }

*RULE4* = SELECT DISTINCT ?sector ?e1 ?e2 ?ep1 ?ep2 ?year2  WHERE {
       ?annual_ets_scenario_by_industrial_sector1 a vocab:annual_ets_scenario_by_industrial_sector ;
         vocab:annual_ets_scenario_by_industrial_sector_employment ?e1 ;
 vocab:annual_ets_scenario_by_industrial_sector_industrial_sector_ref ?sector ;
 vocab:annual_ets_scenario_by_industrial_sector_annual_ets_scenario_ref ?annual_scenario .
       ?annual_ets_scenario_by_industrial_sector2 a vocab:annual_ets_scenario_by_industrial_sector ;
 vocab:annual_ets_scenario_by_industrial_sector_employment ?e2 ;
 vocab:annual_ets_scenario_by_industrial_sector_industrial_sector_ref ?sector ;
 vocab:annual_ets_scenario_by_industrial_sector_annual_ets_scenario_ref ?annual_scenario2  .
       ?sector a vocab:industrial_sector .
       ?annual_scenario a vocab:annual_ets_scenario ;
           vocab:annual_ets_scenario_energy_price ?ep1 ;

```
        vocab:annual_ets_scenario_year ?year1 ;
        vocab:annual_ets_scenario_scenario ?scenario .
   ?annual_scenario2 a vocab:annual_ets_scenario ;
        vocab:annual_ets_scenario_energy_price ?ep2 ;
        vocab:annual_ets_scenario_year ?year2 ;
        vocab:annual_ets_scenario_scenario ?scenario .
   FILTER  (  (?year2 = ?year1 + 5) &&  (?ep2 > ?ep1) && (?e2 < ?e1)  ) }
```